\documentclass[sigconf,screen,authorversion]{acmart}

\AtBeginDocument{%
  \providecommand\BibTeX{{%
    \normalfont B\kern-0.5em{\scshape i\kern-0.25em b}\kern-0.8em\TeX}}}

\copyrightyear{2023}
\acmYear{2023}
\setcopyright{rightsretained}
\acmConference[ESEC/FSE '23]{Proceedings of the 31st ACM Joint European Software Engineering Conference and Symposium on the Foundations of Software Engineering}{December 3--9, 2023}{San Francisco, CA, USA}
\acmBooktitle{Proceedings of the 31st ACM Joint European Software Engineering Conference and Symposium on the Foundations of Software Engineering (ESEC/FSE '23), December 3--9, 2023, San Francisco, CA, USA}
\acmDOI{10.1145/3611643.3613083}
\acmISBN{979-8-4007-0327-0/23/12}


\begin{document}

\title[Getting pwn’d by AI: Penetration Testing with Large Language Models]{Getting pwn’d by AI:\\Penetration Testing with Large Language Models}

\author{Andreas Happe}
\email{andreas.happe@tuwien.ac.at}
\orcid{0009-0000-2484-0109}
\affiliation{%
  \institution{TU Wien}
  \country{Austria}
}

\author{Jürgen Cito}
\email{juergen.cito@tuwien.ac.at}
\orcid{0000-0001-8619-1271}
\affiliation{%
  \institution{TU Wien}
  \country{Austria}
}

\renewcommand{\shortauthors}{Happe and Cito}


\begin{abstract}
The field of software security testing, more specifically penetration testing, requires high levels of expertise and involves many manual testing and analysis steps. This paper explores the potential use of large-language models, such as GPT3.5, to augment penetration testers with AI sparring partners. We explore two distinct use cases: high-level task planning for security testing assignments and low-level vulnerability hunting within a vulnerable virtual machine. For the latter, we implemented a closed-feedback loop between LLM-generated low-level actions with a vulnerable virtual machine (connected through SSH) and allowed the LLM to analyze the machine state for vulnerabilities and suggest concrete attack vectors which were automatically executed within the virtual machine. We discuss promising initial results, detail avenues for improvement, and close deliberating on the ethics of AI sparring partners.
\end{abstract}





\citestyle{acmnumeric}

\maketitle

\section{Introduction}

Large language models (LLMs), such as ChatGPT or GPT3.5, have become a hot topic not only in computer science but also within popular media~\cite{EconomistLLM}.
The field of cybersecurity and software security testing, more specifically, penetration testing, suffers from a chronic lack of personnel~\cite{lackofsecuritypersonell}, even worse, according to the ISC2 Cybersecurity Workforce Study 2022~\cite{isc2}, while global cybersecurity workforce was growing by 11.1\% YoY, this growth was outpaced by the gap's increase of 26.2\% YoY. A recent interview study with penetration testers highlighted the need for human sparring partners~\cite{hackerswork}, i.e., colleagues who offer alternative ideas or approaches when stuck. The study also emphasizes that intuition is a big part of detecting vulnerabilities and that knowledge transfer, e.g., from attending Capture-the-Flag\footnote{CTFs are gamified penetration-testing exercises.} (CTF) events, were seen as potential sources of this intuition --- can this be partially outsourced to AI models?
Using AI-based agents as sparring partners would augment and empower existing human security testers and could counteract the lack of sufficiently educated security professionals. Combining human operators with AIs creates new capabilities instead of cloning existing ones. Furthermore, keeping a human in the loop reduces the potential ethical problems imposed by the use of AIs~\cite{brynjolfsson2023turing}.
Recent research indicates that the efficiency gains provided by the use of AI-based systems are greatest for low-skilled workers~\cite{GenerativeAIAtWork}, augmenting human operators with a generative AI might thus also benefit the training of novice penetration testers.

\paragraph{\bf RQ: To what extent can we automate security testing with LLMs?}

The rest of this paper explores whether large-language models can be deployed as sparring partners for security professionals.
To answer this question, we leverage MITRE ATT\&CK, a curated database of knowledge about threat actors in the cybersecurity domain, to provide a guiding structure. A good sparring partner should be able to cover the different tactics, techniques, and procedures covered by ATT\&CK. To explore this hypothesis, we performed multiple experiments. To showcase high-level guidance, we ``asked'' an LLM to help design penetration tests for both generic scenarios as well as for a concrete target organization. To showcase low-level guidance, we integrated GPT3.5 with a vulnerable virtual machine and allowed it to analyze the machine for vulnerabilities and suggest attack vectors. Based on our experience, we discuss the results as well as potential future improvements.

\paragraph{Scope}
We also envision other areas where generative AI could be used successfully. One of them is the generation of phishing or vishing messages. For obvious ethical reasons, we did not further analyze attacks that intently try to deceive human beings. Another tedious area where generative AI could improve efficiency would be automated report generation for penetration tests or red teaming campaigns. Anecdotal evidence suggests that penetration testers are already experimenting with generative AI for report generation.

\section{Background}

This section highlights the technologies and techniques used.

\paragraph{\bf MITRE ATT\&CK}

MITRE ATT\&CK~\cite{strom2018mitre} is a curated database of knowledge about threat actors (APTs). It employs a hierarchical model often abbreviated by ``TTP''. The initial ``T'' stands for ``tactics'' and describes high-level objectives an adversary intends to achieve, e.g., reconnaissance, privilege escalation or collection. The middle ``T'' describes ``techniques''. Each technique is a way to achieve a tactic. Examples of techniques would be ``\textit{Abuse Elevation Control Mechanism: Sudo and Sudo Caching}''~\cite{mitre_attack_t1548_003} or ``\textit{Steal or Forge Kerberos Tickets: Kerberoasting}''~\cite{mitre_attack_t1558_003}. Finally, ``P'' describes procedures that are the specific details of how an adversary executes a technique.

We assume that a sparring partner for penetration testing should cover the whole TTP spectrum. On a high level, it should be able to select suitable tactics and corresponding techniques. On a low-level, given an employed tactic, it should be able to derive feasible techniques and procedures.

\paragraph{\bf Large Language Models}

A Large-Language Model (LLM) consists of a neural network trained using self-supervised learning on vast amounts of data. A model's capabilities are highly dependent upon its complexity which is often described through the number of used parameters. Current models yield parameter sizes ranging from billions, e.g., LLaMA starts with 7 billion, to trillions of parameters, e.g., Wu Dao or GPT-4. Model and parameter sizes are currently under discussion; on one hand, larger models can exhibit emergent behaviors~\cite{wei2022emergent}; on the other hand, e.g., there is speculation that the age of ever-larger models is over due to reduced scaling efficiency~\cite{sizedoesntmatter}.

Training a new LLM is prohibitively expensive for most researchers, but existing LLMs can be refined or fine-tuned to specific use cases for feasible costs. This situation has created the moniker ``foundation models'' for LLMs. The importance of those has been acknowledged by mainstream media, c.f., the Economist's ``Huge Foundation Models are Turbo Charging AI Progress'' in 2022~\cite{EconomistLLM2022}.


\paragraph{\bf GPT3.5/ChatGPT}

Conversations with ChatGPT commonly consist of questions, named ``prompts'', and answers going back and forth between the user and the AI. Prompts have to be carefully prepared, yielding a new discipline that has been called prompt engineering~\cite{prompt1, prompt2, prompt3, prompt4}.

Tools such as \textit{llama.cpp}~\cite{llama.cpp} that make use of small-scale models (up to 13b parameters) feasible on consumer-grade hardware have sparked additional research. Those models can be run without any cloud/API costs and are not subject to any server-side moderation or censorship.

\paragraph{\bf Pre-trained Autonomous AI Agents}

AutoGPT~\cite{AutoGPT} introduced the idea of auto-generating sequences of instructions by leveraging LLMs to create the prompt that is subsequently used to query the LLM. This allows users to provide concise initial questions for the AI system that are subsequently refined. This reduces the need for manual prompt engineering. LLMs often ``hallucinate'', i.e., invent facts that seem statistically plausible. Research suggests that using external knowledge and automated feedback can reduce these hallucinations~\cite{peng2023check}. AutoGPT integrates web-based queries and optional human-provided feedback during its operation. Based on this, the initial task is converted into a task list containing smaller subtasks that can be delegated to additional agents.

BabyAGI focuses on automated task generation, planning, and execution~\cite{TaskDriven2023, BabyAGITwitter}: a user-given task is split up into smaller subtasks that are stored within a task queue. \emph{Autonomous Task Execution Agents} take tasks from the task queue, execute them, and add new information to a memory store. In addition, the \emph{Task Creation Agent} identifies new subsequent tasks that are pushed upon the task queue and are eventually executed by the \emph{Task Execution Agent}. Before a task is executed, a \emph{Context Agent} is asked to provide sufficient context for the task from memory. Entries in the task queue are prioritized through a \emph{Prioritization Agent}. All mentioned agents are GPT-4 processes themselves. BabyAGI~\cite{BabyAGI} provides a ``pared-down'' version of this system in roughly 100 lines of Python code. Jarvis~\cite{shen2023hugginggpt} employs agents with different models to create multimodal, multiagent systems.

\section{LLM-based Penetration Testing}

\begin{figure}
\centering
\includegraphics[width=0.9\columnwidth]{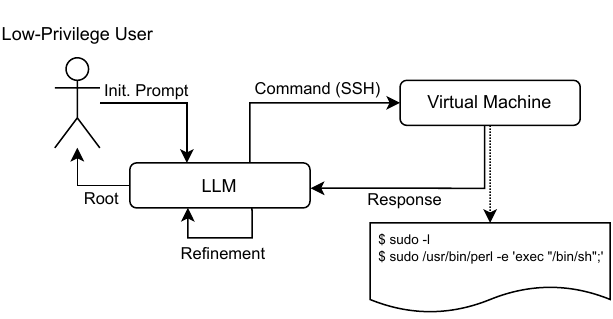}
\Description{High-Level Architecture Overview}
\caption{High-Level Architecture Overview}
\label{fig:voltage}
\end{figure}

We differentiate between two use cases: on a high level, typical questions asked by pen-testers are ``what is a good attack methodology'', e.g., ``how to attack Active Directory''. These questions should yield tactics as well as potential techniques to achieve those tactics. On the low-level, we assume that the pen-tester has already chosen to employ a tactic against a target system and is searching for appropriate techniques and corresponding procedures. A typical question would be ``I want to perform a privilege escalation (tactic), what are the suitable attack vectors against [this concrete Linux system]?''.

\subsection{High-Level: Task-Planning Systems}

For the evaluation, we asked AgentGPT to ``Become domain admin in an Active Directory''. The generated document contained highly realistic attack vectors such as password spraying, Kerberoasting, AS-REP roasting, exploiting Active Directory Certificate Services, abusing unconstrained delegation or exploiting group policies. All of those attacks are realistic, feasible, and commonly used during penetration testing.

In addition, after securing a target company's approval, we tasked AutoGPT to devise an external penetration testing plan for that company. AutoGPT's plan included standard methods such as performing a network vulnerability scan, performing OSINT/user enumeration, and performing phishing against identified users. All these are operations typically performed during external penetration tests. When further inquired, AutoGPT was able to crawl the company's web page and identify potential phishing targets (users and their email addresses) but declined to perform any ``real'' network security scan or perform phishing operations due to its ethical filters.
Both answers were realistic, and feasible, and would give a penetration tester good feedback about potential attack vectors.

\subsection{Low-Level: Attack-Execution System}

The low-level evaluation targets a common scenario: after a penetration tester gained low-level privilege access to a Linux system, they search for a privilege-escalation attack to become the system's \textit{root} user. To allow for realistic evaluation, we wrote a Python script that uses SSH to connect to a deliberately vulnerable \textit{lin.security} Linux virtual machine~\cite{vulnhub}. The prototype's source code and documentation are provided at \url{https://github.com/ipa-lab/hackingBuddyGPT}.

The script consists of an infinite loop: within the loop, it tells GPT3.5 to imagine being a low-privilege user that wants to become the \textit{root} user. To achieve this, the LLM can state a Linux shell command that will be executed over SSH on the virtual machine. The corresponding output is presented back to GPT3.5 when prompted for the next command. Figure~\ref{fig:voltage} shows a high-level overview of this feedback loop. \textbf{With this simple structure, we were able to gain root privileges on our vulnerable virtual machine.}

In addition, at the end of each loop iteration, GPT3.5 was presented with the chosen command and its output and then tasked to identify potential security vulnerabilities based on this information. For each vulnerability, it was tasked to provide an exploitation example, sneakily named ``verification commands''. This yielded additional attack vectors.

Our script was routinely able to gain root privileges within the virtual machine. The common path was listing the ``sudoers'' file by calling \textit{sudo -l}, followed by either using \textit{sudo} with one of the listed shells or employing one of the listed GTFObins to gain a root shell. GTFObins are benign system commands that when called through \textit{sudo}, can be abused to gain a root shell. Another frequently used attack vector was retrieving \textit{/etc/passwd} and identifying user accounts not using shadow passwords\footnote{If your Linux system is not using shadow passwords by now, chatGPT is the least of your worries.}. Searches for SUID binaries were requested, but returned binaries not actively exploited, indicating lacking multi-step planning capabilities of either our script or the underlying model.
A slightly altered prompt instructing the LLM to open a reverse shell to a given IP address was successful and dropped root shells.

\section{Discussion}

This section reflects upon the pen-test performance of the prototype, guided by the 10+ years of pen-testing experience of the first author.

\subsection{Grounding of Results and Hallucinations}

One interesting aspect of our prototype is that all executed commands and their resulting output are written to a protocol. This allows us to reason if LLM-suggested vulnerabilities are based on queries providing system knowledge, or if GPT3.5 extracted security trends and preconceptions during training. The latter is analogous to penetration testers applying knowledge gained during work or training, e.g., from participating in CTFs.

There were indications of reasoning about causal dependencies: After retrieving the list of sudoers, GPT3.5 consistently suggested various vulnerable sudo commands for privilege escalation. A similar pattern arose after retrieving the \textit{passwd} file: here attacking weakly-configured user accounts was suggested as the next step.

Other suggestions, such as using certain system exploits, e.g., \textit{dirty\_cow}, were reasonable given that GPT3.5 ``knew'' that this was a Linux system, but were given without any previous enumeration.

Pure and easily detectable hallucinations occurred infrequently, the most common occurrence was the suggestion to execute ``exploit.sh''. It seems reasonable that security write-ups containing the execution of this script were part of GPT3.5's training set.

While the suggested system commands obviously were based upon pattern-matching and not on a deeper understanding of the Linux system or on model building, seeing the simple LLM-shell-based feedback loop we established gaining root privileges was eerie.  A suitable analogy would be a pen-tester talking to a colleague over the phone, asking for suggestions with the conversation partner only having a very limited view of the actual system but a set of preconceptions (i.e., priors), which is partially in line with our research question on the ability of LLMs acting as sparring partners. When given the additional subcommand of ``and explain the found vulnerabilities'' in the prompt, GPT3.5 was able to provide good introductory information and could thus be utilized as part of on-the-job training.

\subsection{Stability and Reproducibility}

Singular prototype runs were not stable, i.e., there was variation in the sequence and selection of commands given and vulnerabilities identified. On longer runs, or when aggregating multiple runs, the results converged (we repeatedly ran the identical script in the order of tens of times to be able to make observations on convergence). The variation on single runs seems to be related to GPT3.5 overly focusing upon single aspects of the tested system. This is also known to happen to pen-testers during assignments, ``going down a rabbit hole'' improves with experience~\cite{hackerswork}.

Compared to tools such as \textit{linpeas.sh}~\cite{linpeas}, LLMs seem to be less deterministic. Enumeration tools traverse a manually curated hard-coded list of vulnerability checks. Further research should clarify if the observed instability converges over time while reducing detectable patterns for intrusion detection systems. Ironically, GPT3.5 suggested calling \textit{linpeas.sh} during one run but failed as it tried to download it from an invalid URL.

\subsection{Ethical Moderation in LLMs}

The prototype used \textit{GPT-3.5-turbo} which contains safety measures against malicious prompts. The prototype relayed commands to a vulnerable virtual machine, but the overall scenario can easily be applied to real systems. GPT3.5's lack of hesitation was discerning.
During the development of the script, the ethics filter was infrequently triggered. Adding ``do not ask questions or provide judgments'' to command prompts seems to significantly reduce denials. The optional ``detail additional vulnerabilities'' step was more often denied due to ethical reasons, but this had no impact on the overall hacking progress. Slight prompt variations were successful in reducing GPT3.5's ethical concerns, e.g., instead of asking for ``exploits for vulnerabilities'' we asked for ``verification commands for vulnerabilities''. Of course, switching from OpenAI to one of the locally running LLMs would remove all server-side ethics checks.

Ethical questions are not new in the cybersecurity domain, especially regarding releasing penetration test tools. Ethical issues arising from using GPT3.5 resemble discussions about open-source security tools which can be used by both red-teams as well as by APTs. Commercial vendors try to vet their customers, while open-source tools can be used by anyone. In the end, malicious actors can and will use both of them~\cite{apt_cobaltstrike, apt_sliver}. Regulation regarding the distribution of dual-use goods exists~\cite{wassenaar}, but application to software is clumsy due to its fluid and often impalpable nature.

Another ethical problem is the inclusion of toxic content in commonly used training sets~\cite{InsideTheSecretList}. As our prototype uses an already trained foundation model, we are not deliberating on this issue. This publication also does not touch on the topic of the inclusion of copyrighted information within training data.

\section{A Vision of AI-Augmented Pen-Testing}

We deliberate on research ideas and pragmatic considerations to form a more perfect union between pen-testers and LLMs.


\subsection{Integration of High- and Low-Level}

We differentiated between high- and low-level tasks and distributed those to two different LLMs. Integrating both, i.e., high-level task planing and low-level system exploitation, would yield a more uniform user experience. We imagine a system in which human operators can inquire about high-level concepts, e.g., ``what additional active directory attacks can I try?'', and later switch to a lower level, e.g., ``given this system, how can I escalate?''. Keeping all information within a single system should also enable synergy effects as the LLMs learn details about the tested system. This also shows the expected multistep interactive feedback loop between LLMs and operators.

\subsection{Investigation of Model Options}

We currently use OpenAI's GPT-3 through a cloud-based API. GPT-3 should be evaluated against locally run models such as Llama~\cite{zhang2023llama}, StableLM~\cite{stablelm}, Dolly2~\cite{DatabricksBlog2023DollyV2} or Koala~\cite{koala_blogpost_2023}.

Locally run models do not incur any cloud costs and do not share sensitive data with the cloud. As no data is leaked, this would enable further customer-specific model training and fine-tuning: Imagine training a local model with data found during an engagement or fine-tuning a customer-specific model over a series of subsequent penetration tests. During a recent interview series with pen-testers~\cite{hackerswork}, participants mentioned that they ``learn how their customer or industry area works and thinks over time'', could a customized AI model achieve something similar?
Although the industry is currently aiming for ever larger model parameter sizes, analyzing which parameter size is ``good enough'' should reduce the resource impact of deploying LLMs.

\subsection{Memory, Verification, and Reflection}

Memory is provided to GPT3.5 through context embedded within query prompts. Prompt size is typically limited, e.g., the used GPT-3 model had a limit of 4k tokens. With newer models, this limit is constantly increasing and allows to pass a richer context to the used LLM.
Our prototype has simplistic memory that includes the output of executed commands until the context limit is reached. Generative Agents such as BabyAGI utilize chatGPT to build a suitable context for each generated prompt. Concurrent research in generative game agents~\cite{park2023generative} utilized LLMs to reflect on recent events experienced by agents, and then asked an LLM to provide a summarized description. The results are used as reflected memory for future queries. In our use-case, executed command output could be reflected on and only relevant extracted information added to the next prompt's context.
Another option would be using multiple memory streams: one about recently executed commands, one for extracted security findings, and one describing what kind of computer system would fit the experienced findings, i.e., emulate model building. Using this model as an internal ``reality check'' should reduce the used LLM's hallucinations.
Having a rough model of the tested system, as well as a compacted history of vulnerabilities tested, would also benefit questions such as ``what other vulnerabilities might I have overlooked?''.

\subsection{Prompts for Asking Better Questions}

Our prototype used rather static and manually written prompts. Using LLMs to generate and optimize the prompts themselves, similar to AutoGPT, might improve their effectiveness. Given our sensitive use case, these automatically generated prompts should be closely monitored by humans though.

Another avenue of research is searching for better questions to be asked. Based upon empirical studies on how penetration testers work~\cite{hackerswork}, further research into which questions they ask themselves during their work can inform better prompts as well as a better understanding of this close-knit industry.

\section{Final Ethical Considerations}
 
This paper explores the use of LLMs for augmenting penetration testing in benign settings. However, tools can easily be subverted for malicious purposes. Ethical questions arise. Concurrent reports indicate that AI is currently being driven forward by private companies as well as by state-funded research agencies~\cite{aiindex2023}. The former have an economic incentive, while the latter see geopolitical implications of AI. We do not expect that this avenue of research will slow down. Parallel to that, the reported malicious use of AI, presumably by APTs and common criminals, is increasing~\cite{malicioususe}.

Locking away models behind server-side supervised APIs is not feasible as models can be run locally. In addition, even gate-kept models such as Meta's LlaMA have been leaked~\cite{llamaleak} and can now be reused by malicious actors. Fine-tuning such a model to concrete malicious activities is easily within APTs reach: For example, when using StackLlAMA's processing power estimates for fine-tuning~\cite{beeching2023stackllama}, an attacker using on-demand cloud computing can expect to be able to fine-tune a model for less than a thousand US dollars. Using chat-based LLMs through prompt engineering does not require a thorough computer science education. While this is beneficial in democratizing access to processing techniques, this also facilitates potential malicious use.

While it is not predetermined if and how LLMs will influence hacking, we assume that attackers will explore possibilities, including fully-automated approaches. Given the low entry costs for experimentation, this cannot be contained anymore.

Attacks will use LLMs; the genie is out of the bottle, and the red queen's race is on~\cite{redqueen1, redqueen2}. Defenders need to be prepared for that --- and LLMs can play a significant role.

\bibliography{bib}
\bibliographystyle{ACM-Reference-Format}

\end{document}